\documentclass[10pt,twocolumn,letterpaper]{article}

\usepackage{iccv}              
\usepackage{graphicx}
\usepackage{verbatim}
\usepackage{multirow}
\usepackage{subcaption} 
\usepackage{xcolor}
\usepackage{float}
\usepackage{colortbl}  
\usepackage{tikz}      
\graphicspath{{figs/}}
\makeatletter
\def\input@path{{tbls/}} 
\definecolor{iccvblue}{rgb}{0.21,0.49,0.74}
\usepackage[pagebackref,breaklinks,colorlinks,allcolors=iccvblue]{hyperref}

\def\paperID{5310} 
\def\confName{ICCV}
\def\confYear{2025}

\title{VisHall3D: Monocular Semantic Scene Completion from Reconstructing the Visible Regions to Hallucinating the Invisible Regions
}

\author{Haoang Lu
\quad
Yuanqi Su\thanks{Corresponding Author. Email: yuanqisu@mail.xjtu.edu.cn}
\quad
Xiaoning Zhang
\quad
Longjun Gao
\quad
Yu Xue
\quad
Le Wang\\
Xi'an Jiaotong University, China\\
\url{https://github.com/luang321/vishall3d}}

\begin{document}
\newcommand{\colorboxsquare}[2]{%
\ensuremath{
    \stackrel{\rotatebox{90}{\footnotesize #2}}{%
        \tikz[baseline=(char.base), xshift=1mm]\node[fill=#1, minimum width=2.5mm, minimum height=2.5mm, inner sep=0pt] (char) {};%
    }
    }
}

\maketitle

\begin{abstract}
This paper introduces VisHall3D, a novel two-stage framework for monocular semantic scene completion that aims to address the issues of feature entanglement and geometric inconsistency prevalent in existing methods. VisHall3D decomposes the scene completion task into two stages: reconstructing the visible regions (vision) and inferring the invisible regions (hallucination). In the first stage, VisFrontierNet, a visibility-aware projection module, is introduced to accurately trace the visual frontier while preserving fine-grained details. In the second stage, OcclusionMAE, a hallucination network, is employed to generate plausible geometries for the invisible regions using a noise injection mechanism. By decoupling scene completion into these two distinct stages, VisHall3D effectively mitigates feature entanglement and geometric inconsistency, leading to significantly improved reconstruction quality.

The effectiveness of VisHall3D is validated through extensive experiments on two challenging benchmarks: SemanticKITTI and SSCBench-KITTI-360. VisHall3D achieves state-of-the-art performance, outperforming previous methods by a significant margin and paves the way for more accurate and reliable scene understanding in autonomous driving and other applications.
\end{abstract}

\begin{figure*}[!htbp]
  \centering
    \includegraphics[width=\linewidth]{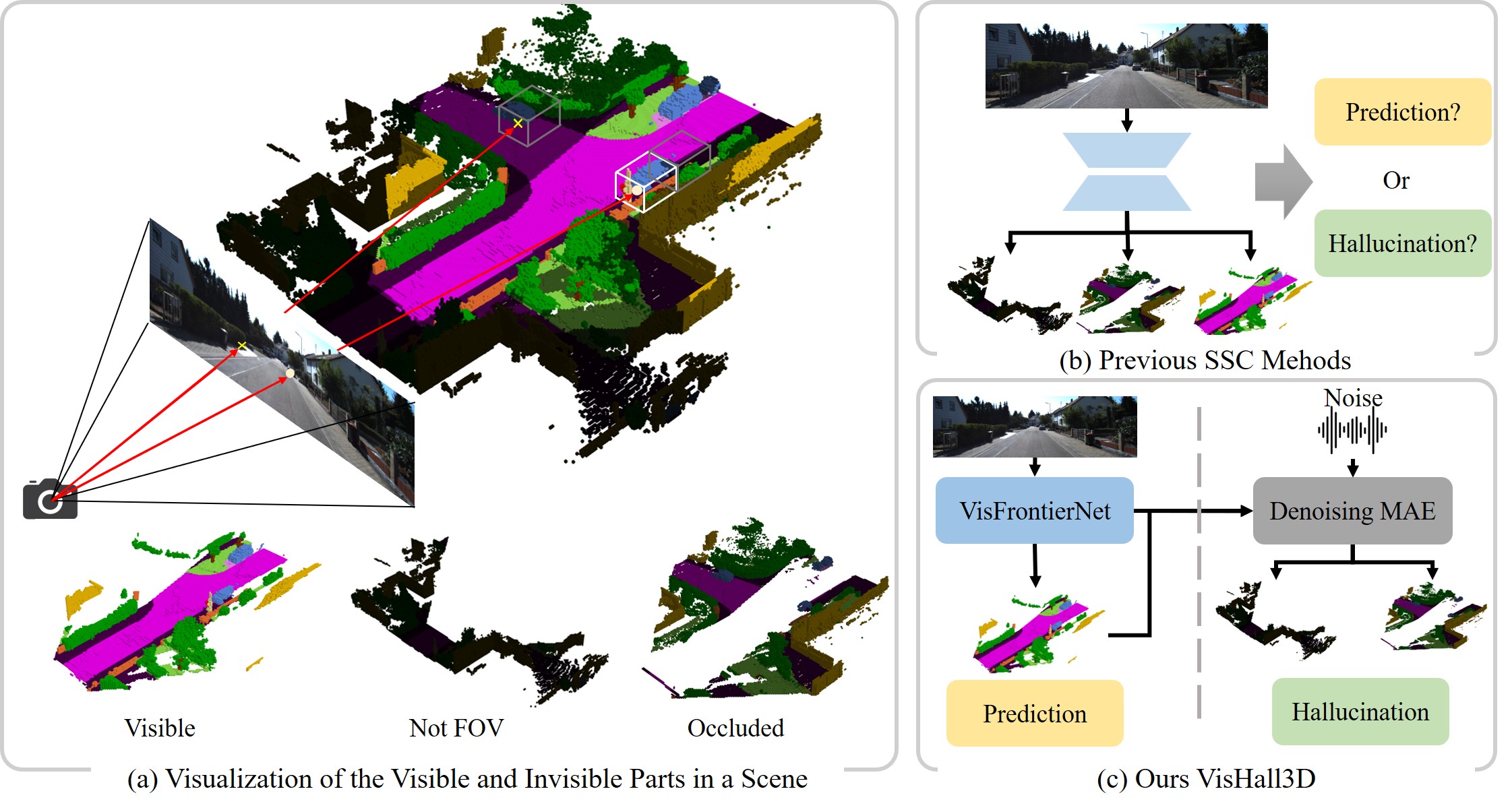}
  \caption{Visualization of the decoupling of prediction for visible regions and hallucination for invisible regions: (a) Division of visible and invisible areas: OOV (Out-of-view) voxels are in black, and occluded FOV (Field-of-View) voxels are in dark; (b) Pipeline of current mainstream methods; (c) Pipeline of our approach and its decoupling strategy for prediction and hallucination.}
  \label{fig:mot}
\end{figure*}
\section{Introduction}
\label{sec:intro}
Monocular Semantic Scene Completion (Monocular SSC)~\cite{cao2022monoscene,miao2023occdepth,yao2023ndc,li2023voxformer,CGFormer} has emerged as a promising solution for enabling autonomous vehicles to perceive and understand their surroundings, as it can reconstruct complete 3D scenes using only single RGB images. This cost-effective and flexible approach has the potential to revolutionize 3D perception in autonomous driving. However, existing Monocular SSC methods still suffer from two major challenges: feature entanglement and geometric inconsistency.

These issues stem from the inherent distinction between visible and invisible regions in the 3D scene. The visible regions, or the visual frontier, correspond to the surface points of the 3D scene captured in the input image. When estimating 3D occupancy grids, the ground truth is typically obtained by accumulating data from multiple adjacent Lidar frames, inevitably including some voxels that are not visible in the current image. As a result, the occupancy estimation task faces two distinct challenges: tracing the visual frontier and generating the unseen voxels, as shown in Fig.\ref{fig:mot}.

Existing methods, such as MonoScene~\cite{cao2022monoscene}, Occdepth~\cite{miao2023occdepth}, and NDC-Scene~\cite{yao2023ndc}, among others~\cite{li2023voxformer,zheng2024monoocc,wang2024h2gformer,jiang2024symphonize,CGFormer}, have made significant progress in Monocular SSC. However, by treating SSC as a single-stage task, they fail to recognize the inherent distinction between visible and invisible regions, leading to feature entanglement and geometric inconsistency. Feature entanglement occurs when the features learned for visible and invisible regions are mixed together, while geometric inconsistency arises when the reconstructed visible and invisible regions are misaligned or have conflicting structures.

To address these challenges, we propose VisHall3D, a novel two-stage framework that explicitly separates the tasks of reconstructing visible regions (vision) and inferring invisible regions (hallucination). In the first stage, VisHall3D uses VisFrontierNet, a visibility-aware projection module, to accurately trace the visual frontier while preserving fine-grained details. By explicitly modeling the boundary between visible and invisible regions, VisFrontierNet helps to mitigate feature entanglement. In the second stage, VisHall3D employs {OcclusionMAE}, a hallucination network that generates plausible geometries for invisible regions using a noise injection mechanism. By decomposing SSC into these two distinct stages, VisHall3D effectively addresses feature entanglement and geometric inconsistency, leading to significantly improved reconstruction quality, as shown in Fig.\ref{fig:mot}.

We validate the effectiveness of VisHall3D through extensive experiments on SemanticKITTI~\cite{behley2019semantickitti} and SSCBench-KITTI-360~\cite{li2024sscbench,liao2022kitti}, achieving state-of-the-art performance. Our main contributions are:
\begin{itemize}
\item A novel two-stage framework, VisHall3D, for Monocular SSC that explicitly separates vision and hallucination processes to mitigate feature entanglement and geometric inconsistency.
\item VisFrontierNet, a visibility-aware projection module that accurately traces the visual frontier by modeling the boundary between visible and invisible regions.
\item OcclusionMAE, a hallucination network that generates plausible geometries for invisible regions using a noise injection mechanism.
\item VisHall3D sets a new standard for Monocular SSC, paving the way for more accurate and reliable scene understanding in various applications.
\end{itemize}

\begin{figure*}[htbp]
  \centering
  \includegraphics[width=\linewidth]{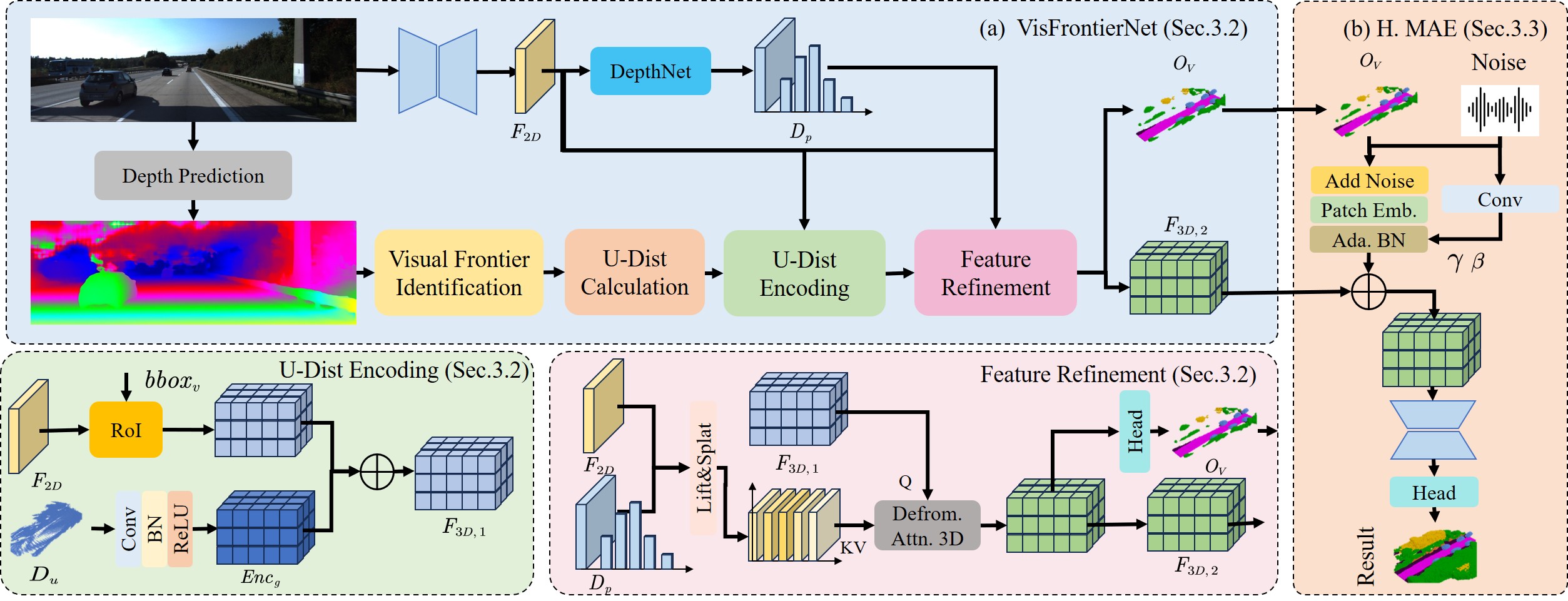}
  \caption{Overview of our method and the structure of each module. (a) The pipeline of our proposed VisFrontierNet, including unsigned distance (uDistance) Encoding and Feature Refinement. (b) The detailed architecture of the OcclusionMAE. The unsigned distance is calculated for the identified visual frontier which is then encoded and combined with image feature for the 3D representation.}
  \label{fig:pipeline}
\end{figure*}
\section{Related Work}
\subsection{Semantic Scene Completion}
3D Semantic Scene Completion aims to generate dense 3D semantic voxel grids from incomplete observations, as first defined in SSCNet~\cite{song2017semantic}. Early SSC works typically relied on geometric inputs such as LiDAR~\cite{cheng2021s3cnet,yan2021sparse,rist2021semantic} or depth maps~\cite{song2017semantic,li2020anisotropic,li2019depth}. However, recent studies have explored reconstructing entire SSC scenes using visual-only inputs, starting from multi-view methods~\cite{li2022bevformer,huang2023tri}, then progressing to stereo-based approaches~\cite{li2023bridging}, and finally to monocular frameworks~\cite{cao2022monoscene,yao2023ndc}.

Among multi-view methods, BEVFormer~\cite{li2022bevformer} and TPVFormer~\cite{huang2023tri} leverage diverse 3D feature representations to reduce computational overhead and enhance network capability. FB-OCC~\cite{li2023fb} and OccTransformer~\cite{liu2024occtransformer}, on the other hand, focus on improving the transition from 2D to 3D.

Monoscene~\cite{cao2022monoscene} pioneered single-image SSC, while subsequent works introduced NDC coordinates~\cite{yao2023ndc}, depth prediction~\cite{li2023voxformer}, horizontal direction emphasis~\cite{wang2024h2gformer}, and voxel-to-graph structures~\cite{yao2023depthssc} to optimize 2D-to-3D transformation. The latest advancements, Symphonies~\cite{jiang2024symphonize} and CGFormer~\cite{CGFormer}, refine the approach with object-aware and context-aware optimizations. However, these methods still suffer from feature entanglement and geometric inconsistency due to the inherent ambiguity in monocular 3D reconstruction.

\subsection{3D From a Single Image}
3D reconstruction from a single image is an ill-posed problem due to the lack of explicit depth information. Early monocular 3D tasks focused on reconstructing coarse information within the visible region by extending 2D methods~\cite{zhou2019objects,luo2021m3dssd}, exploiting geometric knowledge of objects~\cite{lavreniuk2024spidepth,zhang2023monodetr,zhang2021objects,li2022diversity}, or treating it as a Perspective-n-Points (PnP) problem~\cite{chen2022epro,liu2021autoshape}.

The emergence of monocular semantic scene completion (SSC)~\cite{cao2022monoscene,zheng2024monoocc,wang2024not} has expanded the scope of monocular 3D tasks to reconstructing both visible and invisible regions with detailed shapes and semantics. This requires models to possess not only visual perception capabilities but also the ability to hallucinate missing information. However, existing methods often struggle to effectively handle the distinction between visible and invisible regions, leading to suboptimal results.

\subsection{Multi-stage Refinement}
Multi-stage refinement strategy has been widely adopted in computer vision, particularly in object detection~\cite{cai2019cascade,carion2020end,zhu2020deformable,sun2021sparse}, to refine proposals and enhance accuracy. Inspired by this, we propose to decouple the tasks of visual perception and hallucination in monocular SSC. We first generate a coarse prediction of the visible regions, and then leverage Hallucinating MAE~\cite{he2022masked} to hallucinate and optimize the entire scene.

\section{VisHall3D: VisFrontierNet and OcclusionMAE}

\begin{figure}[htbp]
  \centering
  \includegraphics[width=1\linewidth]{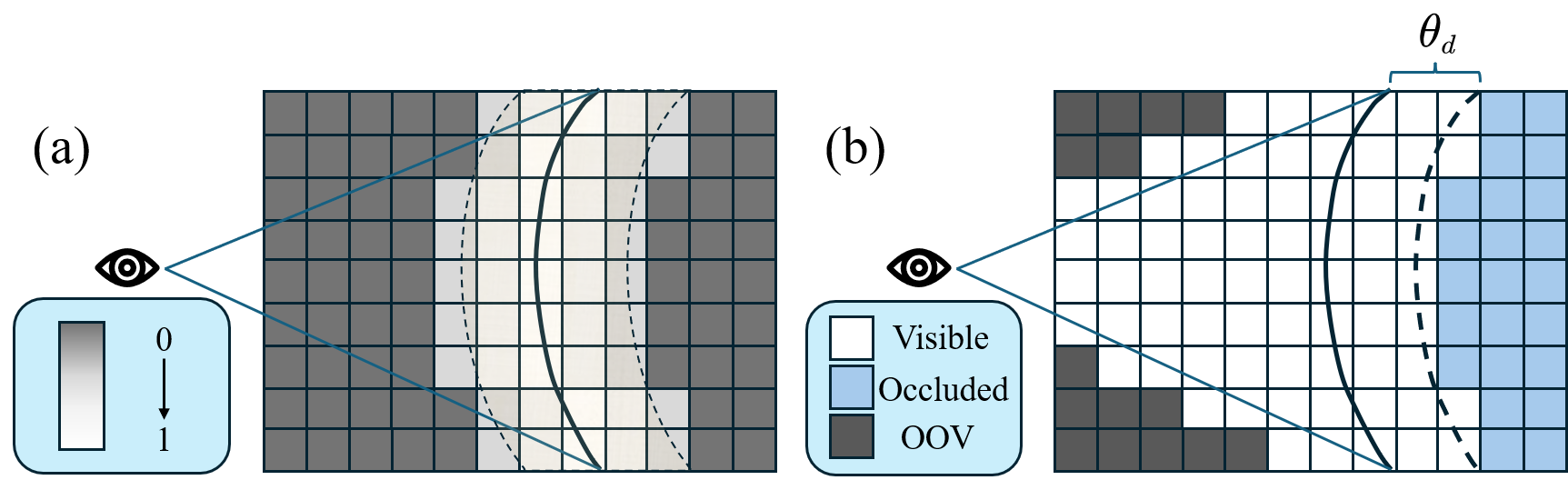}
   \caption{Illustration of the visual frontier from side view. The solid line gives the estimated depth.}
   \label{fig:od}
\end{figure}

We propose VisHall3D, a two-stage framework for generating a 3D occupancy grid from a single image: (1) refinement of the visual frontier by \emph{VisFrontierNet} and (2) completion of the invisible voxels through \emph{OcclusionMAE}, as shown in Fig.\ref{fig:pipeline}. The visual frontier refers to the set of voxels directly visible from the camera's viewpoint, corresponding to the surface points of the 3D scene captured in the input image, as shown in Fig.\ref{fig:od}.

In Sec.\ref{sec:overview}, we provide an overview of the proposed model, including its overall architecture and key components. We then delve into the details of \emph{VisFrontierNet} in Sec.\ref{sec:VisFrontierNet}, explaining its role in refining the visual frontier. Subsequently, in Sec.\ref{sec:OcclusionMAE}, we present \emph{OcclusionMAE} and its approach to completing the invisible voxels. Each subsection offers a comprehensive explanation of the respective modules and their roles within the framework.

\subsection{Method Overview}\label{sec:overview}
In our approach, we divide the 3D occupancy grid voxels into visible, occluded, and out-of-view categories based on visibility (Fig.\ref{fig:mot}). Visible voxels are directly observed; occluded voxels are hidden by obstructions; out-of-view voxels lie beyond the camera's field of view. 

Our proposed method seamlessly integrates feature lifting and 3D generation to reconstruct accurate 3D occupancy from a single input image. 

\textbf{Estimating Visible Voxel Occupancy:}
For visible voxels, we leverage the direct correspondence between the 2D image pixels and their corresponding 3D coordinates in the occupancy grid. Given the estimated depth map $D$, we determine the visible frontier in the 3D space and identify the voxels in front of the frontier as visible voxels.

To capture the frontier’s geometry, we compute a truncated unsigned distance map  $D_u$ representing voxel distances to the frontier. This map is encoded using 3D convolutions and fused with image features projected into 3D space using voxel-projected bounding boxes $bbox_v$ and RoI pooling, injecting visual information into the 3D volume.

We then refine the visual frontier by lifting 2D features into 3D using a depth probability map $D_p$ and scattering them onto a discretized view frustum. Finally, 3D deformable attention refines frontier features, focusing the network on informative regions for better reconstruction.




\textbf{Inferring Occluded and Out-of-View Voxel Occupancy:}
For occluded and out-of-view voxels, the direct visual evidence is not available, requiring a more sophisticated generation process. We propose OcclusionMAE, a denoising Masked Autoencoder that generates the occupancy for the entire set of voxels by taking the noisy prediction on the visible voxels from VisFrontierNet as input.

The denoising procedure in OcclusionMAE involves adding noise to the visible voxel predictions and injecting the noise level into the network via adaptive batch normalization. OcclusionMAE employs a 3D U-Net architecture to process the modulated features and generate the final occupancy grid, effectively aligning the results with the image semantics while accounting for the noisy nature of the visible voxel predictions.

\subsection{Visual Frontier Propagation via VisFrontierNet}\label{sec:VisFrontierNet}
VisFrontierNet is designed to estimate the occupancy of visible voxels by leveraging the correspondence between 2D image pixels and their 3D coordinates in the occupancy grid. Given the 3D grids $G$, our goal is to predict the label of each voxel $v \in G$ from an input image. The main steps of VisFrontierNet include: (1) identifying the visual frontier, (2) representing and encoding it, and (3) refining its feature.

We first transform the coordinates of each voxel into the camera frame; then project them onto the image plane, obtaining the corresponding coordinates $(x_v, y_v, d_v)$ for each voxel $v$. Let $\pi$ denote the projection process, it maps each voxel $v$ to a corresponding 2D coordinate $(x_v, y_v)$ on the image plane.
\begin{equation}
(x_v,y_v,d_v)= \pi(v)
\end{equation}
Here, we also obtain the depth value $d_v$ for each voxel. 

\textbf{Visual Frontier Identification.} The estimated depth map $D$ gives a visual frontier that determines the visible and occluded voxels. The voxels in front of the visual frontier are considered visible,as shown in Fig.~\ref{fig:od}.
\begin{equation}\label{eq:visible}
V:= \{v|v\in G, d_v < D(x_v,y_v) + \theta_d\}
\end{equation}
By comparing the projected depth $d_v$ with the estimated depth of the projected point $D(x_v,y_v)$, we determine whether the voxel lies in front of or behind the visual frontier. A relaxation factor $\theta_d$ is added to accommodate potential errors in depth estimation. The filtered visible voxels $V$ allow our network to focus on estimating the correct lifting from the 2D image plane into the 3D space. 

\textbf{Unsigned Distance Function and Encoding.} To capture the geometric characteristics of the visual frontier, we introduce the truncated unsigned distance map ($D_u$). It measures the unsigned distance $dist_v=|d_v-D(x_v,y_v)|$ from each voxel to the visual frontier, and is defined as:
\begin{equation}
D_u = \begin{cases}
2-2\sigma\left(\gamma* dist_v\right), & dist_v<\theta \\
0, & \text{Otherwise}
\end{cases}
\label{eq:score}
\end{equation}
where $\sigma$ is the sigmoid function, $\gamma$ is a controlling factor set to 10 for sharp decaying, and $\theta$ controls the truncated region (set to 1), as shown in Fig.\ref{fig:od} (a).
\begin{figure}[htbp]
 \centering
\includegraphics[width=1\linewidth]{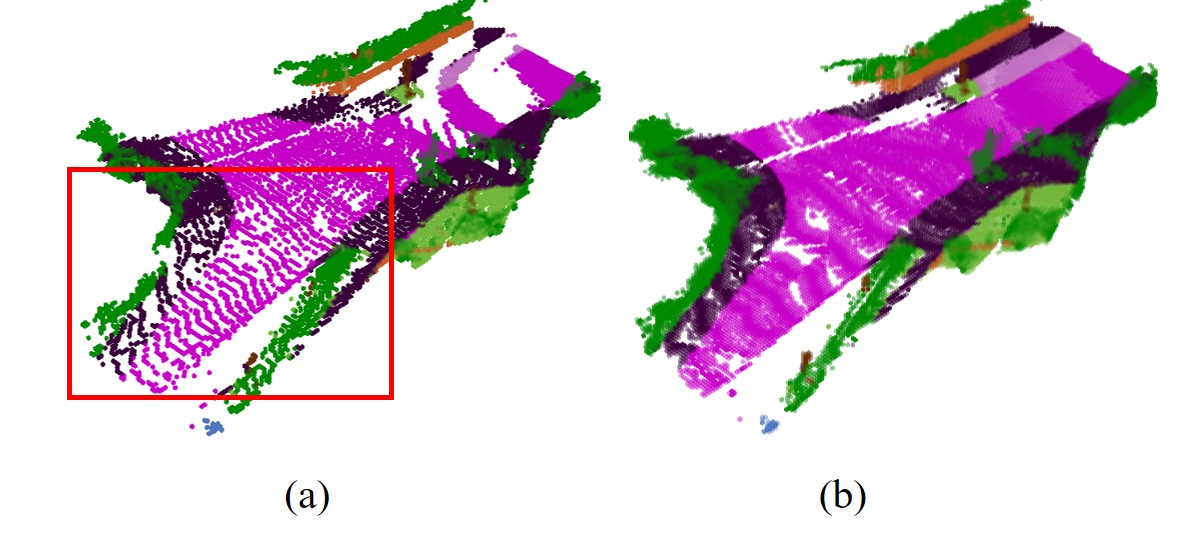}
   \caption{Visual comparison of Hard Lifting and our Unsigned Distance approach. (a) Illustration of Hard Lifting. (b) Geometric information captured by our Unsigned Distance approach.}
   \label{fig:lift}
\end{figure}

Through this approach, we address the challenge of sparse voxels in distant regions, a limitation inherent in the hard lifting of geometric information adopted by previous methods~\cite{li2023voxformer,jiang2024symphonize}, where each pixel corresponds to a single voxel, as illustrated in the Fig.~\ref{fig:lift}.

The unsigned distance map $D_u$ is then fed into a 3D convolutional module for geometric encoding $\text{Enc}_g$, which captures the geometric properties of the visual frontier. To inject the image features into the 3D space, we combine the geometric encoding with the image feature for each voxel using RoI pooling~\cite{girshick2015fast}:
\begin{equation}
F_{3D,1}(v) = \text{Enc}_g(v) \oplus \text{RoI}(F_{2D}, bbox_v)
\end{equation}
where $F_{2D}$ denotes the image feature map, and $bbox_v$ is the projected bounding box for voxel $v$. In practice, We use ResNet-50~\cite{he2016deep} as the backbone for extracting multi-scale features from the input image. Similar to Symphonies~\cite{jiang2024symphonize}, we utilize MaskDINO~\cite{li2023mask}’s neck to combine multi-scale image features into a single one $F_{2D}$.
\begin{figure*}[!htbp]
  \centering
  \includegraphics[width=0.9\linewidth]{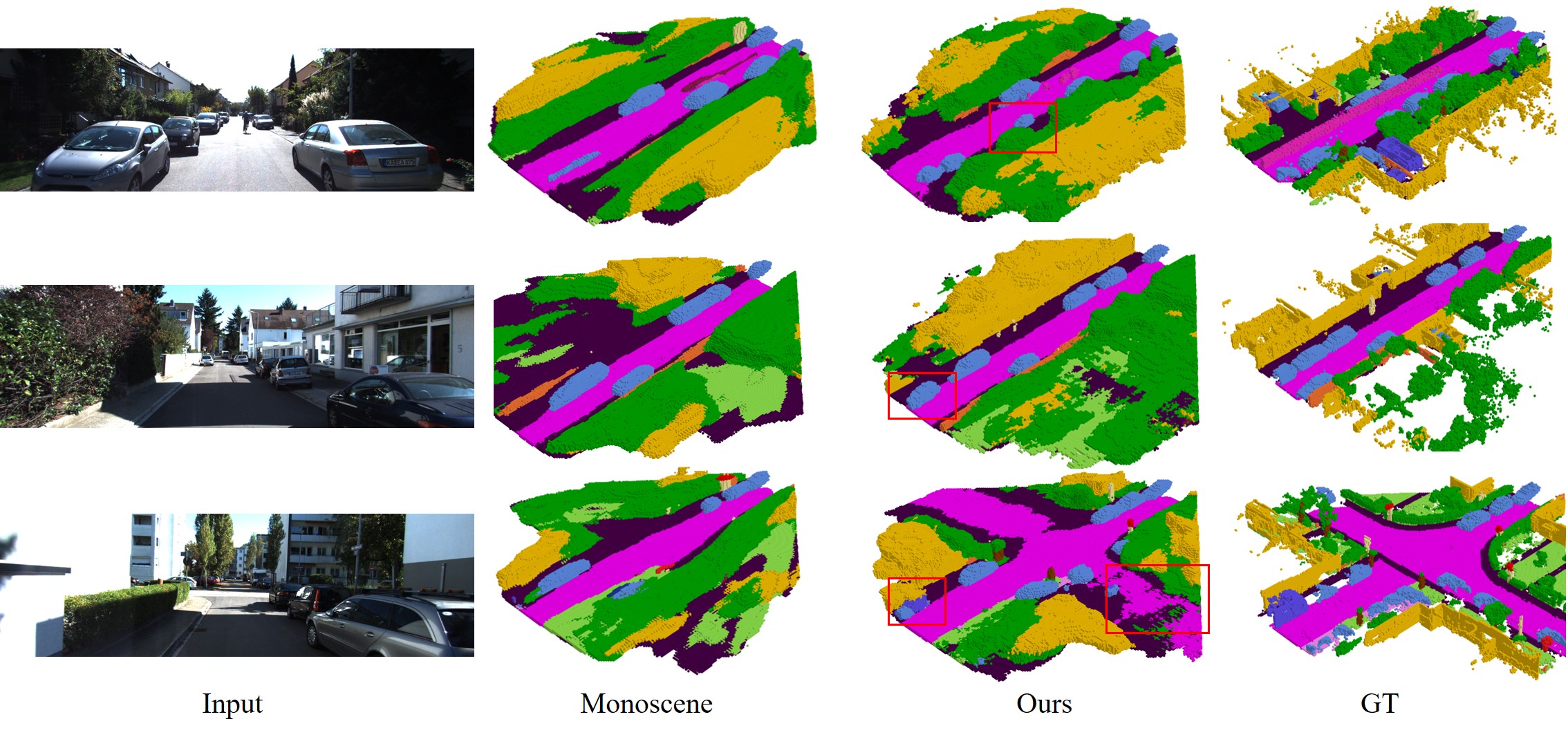}
  \caption{Qualitative visual comparison  with Monoscene\cite{cao2022monoscene} and Ground Truth (GT) on the SemanticKITTI\cite{behley2019semantickitti} dataset.}
  \label{fig:visualization}
\end{figure*}

\textbf{Feature Refinement for Visual Frontier.} After encoding both the geometric and visual features for the frontier, a refinement procedure is utilized to modify the representation for visual frontier accoding to image features. We lift the 2D features $F_{2D}$ into the 3D space using a depth probability map $D_p$, which represents the likelihood of each pixel belonging to different depth ranges. The lifted features are then scattered onto a discretized grid on the view frustum~\cite{philion2020lift}, as shown in Fig.\ref{fig:pipeline}.

To refine the features for the visual frontier, we employ 3D deformable attention~\cite{zhu2020deformable,li2023dfa3d}. It learns to dynamically adjust the receptive field and sampling locations based on the input features, enabling the network to focus on the most informative regions and adapt to the specific characteristics of the visual frontier, producing refined 3D features $F_{3D,2}$.
\begin{equation}
F_{3D,2} \gets \text{Deform3D}(F_{3D,1},F_{2D},D_p)
\label{ff}
\end{equation}

\subsection{Generation via Denoising MAE}\label{sec:OcclusionMAE} 
OcclusionMAE is a denoising Masked Autoencoder designed to generate the occupancy for the entire set of voxels, taking the noisy prediction $O_V$ on the visible voxels from VisFrontierNet as input.

\textbf{Noise Adding to Visible Voxel Predictions.} Considering the noisy nature of the predicted occupancy on visible voxels, we introduce a denoising procedure. For each visible voxel in $V$, we randomly assign it a value from its neighboring voxels within a horizontal range $R_h$ and a depth range $R_d$. Here, considering the spatial resolution of the scene (256 voxels in the horizontal and depth directions, but only 32 voxels in the vertical direction), we focus on adding noise in the horizontal and depth directions while leaving the vertical direction unchanged. The noise adding procedure is described as follows.
\begin{equation}
\tilde{O}_V = \text{AddNoise}(O_V,t(R_h,R_d))
\end{equation}
Where, $t$ is the noise level. It is important to note that the added noise does not alter the original semantics (e.g., changing a vehicle to a person) but only perturbs geometric information (e.g., adjusting voxel positions). This is because we do not want the network to generate non-existent classes.

\textbf{Noise Level Injecting via Adaptive Batch Normalization.} The noise level $t$ is injected into the network using adaptive batch normalization~\cite{huang2017arbitrary}. Let $F$ denote the features extracted from $\tilde{O}_V$ through patch embedding and 3D convolution. The adaptive batch normalization works as follows:
\begin{equation}
ada. BN(F(v)) = \gamma \cdot \frac{F(v) - \mu}{\sqrt{\delta^2 + \epsilon}} + \beta
\label{adabn}
\end{equation}
where $\mu$ and $\delta^2$ are the mean and variance of $F$, respectively. The features from the noise level $t$ are used to generate the parameters $\gamma$ and $\beta$. This allows the network to adapt to different noise levels and generate more accurate occupancy predictions.
\definecolor{road}{RGB}{255, 0, 255}
\definecolor{sidewalk}{RGB}{75, 0, 75}
\definecolor{parking}{RGB}{255, 150, 255}
\definecolor{other-grnd.}{RGB}{175, 0, 75}
\definecolor{building}{RGB}{255, 200, 0}
\definecolor{car}{RGB}{100, 150, 245}
\definecolor{truck}{RGB}{80, 30, 180}
\definecolor{bicycle}{RGB}{100, 230, 245}
\definecolor{motorcycle}{RGB}{30, 60, 150}
\definecolor{other-veh.}{RGB}{100, 80, 250}
\definecolor{vegetation}{RGB}{0, 175, 0}
\definecolor{trunk}{RGB}{135, 60, 0}
\definecolor{terrain}{RGB}{150, 240, 80}
\definecolor{person}{RGB}{255, 30, 30}
\definecolor{bicyclist}{RGB}{255, 40, 200}
\definecolor{motorcyclist}{RGB}{150, 30, 90}
\definecolor{fence}{RGB}{255, 120, 50}
\definecolor{pole}{RGB}{255, 240, 150}
\definecolor{traf.-sign}{RGB}{255, 0, 0}

\begin{table*}[htbp]
    \footnotesize
    \setlength{\tabcolsep}{1pt}
    \renewcommand{\arraystretch}{1.2}
    \centering 
    \begin{tabular}{c|c|cc|ccccccccccccccccccc}
        \toprule
        \textbf{Method} & \textbf{Date} & \textbf{IoU$\uparrow$} & \textbf{mIoU$\uparrow$} & \colorboxsquare{road}{road} & \colorboxsquare{sidewalk}{sidewalk} & \colorboxsquare{parking}{parking} & \colorboxsquare{other-grnd.}{other-grnd.} & \colorboxsquare{building}{building} & \colorboxsquare{car}{car} & \colorboxsquare{truck}{truck} & \colorboxsquare{bicycle}{bicycle} & \colorboxsquare{motorcycle}{motorcycle} & \colorboxsquare{other-veh.}{other-veh.} & \colorboxsquare{vegetation}{vegetation} & \colorboxsquare{trunk}{trunk} & \colorboxsquare{terrain}{terrain} & \colorboxsquare{person}{person} & \colorboxsquare{bicyclist}{bicyclist} & \colorboxsquare{motorcyclist}{motorcyclist} & \colorboxsquare{fence}{fence} & \colorboxsquare{pole}{pole} & \colorboxsquare{traf.-sign}{traf.-sign} \\ 
        \midrule
        \multicolumn{22}{@{}l}{\quad\textbf{Stereo camera-based methods}} \\ \hline
        StereoScene\cite{li2023bridging} & IJCAI2024 & 43.34 & 15.36 & 61.90 & 31.20 & 30.70 & 10.70 & 24.20 & 22.80 & 2.80 & 3.40 & 2.40 & 6.10 & 23.80 & 8.40 & 27.00 & 2.90 & 2.20 & 0.50 & 16.50 & 7.00 & 7.20 \\
        \hline
        \multicolumn{22}{@{}l}{\quad\textbf{Monocular temporal methods}} \\
        \hline
        VoxFormer-T\cite{li2023voxformer} & CVPR2023 & 43.21 & 13.41 & 54.10 & 26.90 & 25.10 & 7.30 & 23.50 & 21.70 & 3.60 & 1.90 & 1.60 & 4.10 & 24.40 & 8.10 & 24.20 & 1.60 & 1.10 & 0.00 & 13.10 & 6.60 & 5.70 \\
        HASSC-T\cite{wang2024not} & CVPR2024 & 42.87 & 14.38 & 55.30 & 29.60 & 25.90 & 11.30 & 23.10 & 23.00 & 2.90 & 1.90 & 1.50 & 4.90 & 24.80 & 9.80 & 26.50 & 1.40 & 3.00 & 0.00 & 14.30 & 7.00 & 7.10 \\
        HTCL\cite{li2024hierarchical} & ECCV2024 & 44.23 & 17.09 & 64.40 & 34.80 & 33.80 & 12.40 & 25.90 & 27.30 & 5.70 & 1.80 & 2.20 & 5.40 & 25.30 & 10.80 & 31.20 & 1.10 & 3.10 & 0.90 & 21.10 & 9.00 & 8.30 \\
        H2GFormer-T\cite{wang2024h2gformer} & AAAI2024 & 43.52 & 14.60 & 57.90 & 30.40 & 30.00 & 6.90 & 24.00 & 23.70 & 5.20 & 0.60 & 1.20 & 5.00 & 25.20 & 10.70 & 25.80 & 1.10 & 0.10 & 0.00 & 14.60 & 7.50 & 9.30 \\
        \hline
        \multicolumn{22}{@{}l}{\quad\textbf{Monocular single-frame methods}} \\
        \hline
        MonoScene\cite{cao2022monoscene}& CVPR2023 & 34.16 & 11.08 & 54.70 & 27.10 & 24.80 & 5.70 & 14.40 & 18.80 & 3.30 & 0.50 & 0.70 & 4.40 & 14.90 & 2.40 & 19.50 & 1.00 & 1.40 & 0.40 & 11.10 & 3.30 & 2.10 \\
        VoxFormer-S\cite{li2023voxformer} & CVPR2023 & 42.95 & 12.20 & 53.90 & 25.30 & 21.10 & 5.60 & 19.80 & 20.80 & 3.50 & 2.60 & 0.70 & 3.70 & 22.40 & 7.50 & 21.30 & 1.40 & 2.60 & 0.20 & 11.10 & 5.10 & 4.90 \\
        TPVFormer\cite{huang2023tri} & CVPR2023 & 34.25 & 11.26 & 55.10 & 27.20 & 27.40 & 6.50 & 14.80 & 19.20 & 3.70 & 1.00 & 0.50 & 2.30 & 13.90 & 2.60 & 20.40 & 1.10 & 2.40 & 0.30 & 11.00 & 2.90 & 1.50 \\
        SurroundOcc\cite{wei2023surroundocc} & ICCV2023 & 34.72 & 11.86 & 56.90 & 28.30 & 30.20 & 6.80 & 15.20 & 20.60 & 1.40 & 1.60 & 1.20 & 4.40 & 14.90 & 3.40 & 19.30 & 1.40 & 2.00 & 0.10 & 11.30 & 3.90 & 2.40 \\
        OccFormer\cite{zhang2023occformer} & ICCV2023 & 34.53 & 12.32 & 55.90 & 30.30 & 31.50 & 6.50 & 15.70 & 21.60 & 1.20 & 1.50 & 1.70 & 3.20 & 16.80 & 3.90 & 21.30 & 2.20 & 1.10 & 0.20 & 11.90 & 3.80 & 3.70 \\
        IAMSSC\cite{xiao2024instance} & T-ITS2024 & 43.74 & 12.37 & 54.00 & 25.50 & 24.70 & 6.90 & 19.20 & 21.30 & 3.80 & 1.10 & 0.60 & 3.90 & 22.70 & 5.80 & 19.40 & 1.50 & 2.90 & 0.50 & 11.90 & 5.30 & 4.10 \\
        DepthSSC\cite{yao2023depthssc} & arXiV2024 & 44.58 & 13.11 & 55.64 & 27.25 & 25.72 & 5.78 & 20.46 & 21.94 & 3.74 & 1.35 & 0.98 & 4.17 & 23.37 & 7.64 & 21.56 & 1.34 & 2.79 & 0.28 & 12.94 & 5.87 & 6.23 \\
        HASSC-S\cite{wang2024not} & CVPR2024 & 43.40 & 13.34 & - & - & - & - & - & - & - & - & - & - & - & - & - & - & - & - & - & - & - \\
        Symphonize\cite{jiang2024symphonize} & CVPR2024 & 42.19 & 15.04 & 58.40 & 29.30 & 26.90 & 11.70 & 24.70 & 23.60 & 3.20 & 3.60 & \underline{2.60} & 5.60 & 24.20 & 10.00 & 23.10 & \textbf{3.20} & 1.90 & \textbf{2.00} & 16.10 & 7.70 & 8.00 \\
        H2GFormer-S\cite{wang2024h2gformer} & AAAI2024 & 44.20 & 13.72 & 56.40 & 28.60 & 26.50 & 4.90 & 22.80 & 23.40 & 4.80 & 0.80 & 0.90 & 4.10 & 24.60 & 9.10 & 23.80 & 1.20 & 2.50 & 0.10 & 13.30 & 6.40 & 6.30 \\
        MonoOcc-L\cite{zheng2024monoocc}& ICRA2024 & - & 15.63 & 59.10 & 30.90 & 27.10 & 9.80 & 22.90 & 23.90 & \underline{7.20} & \textbf{4.50} & 2.40 & \textbf{7.70} & \underline{25.00} & 9.80 & 26.10 & \underline{2.80} & \underline{4.70} & 0.60 & 16.90 & 7.30 & 8.40 \\
        CGFormer\cite{CGFormer} & NIPS2024 & 44.41 & 16.63 & \underline{64.30} & \textbf{34.20} & \textbf{34.10} & \underline{12.10} & \underline{25.80} & \underline{26.10} & 4.30 & \underline{3.70} & 1.30 & 2.70 & 24.50 & \underline{11.20} &\textbf{29.30} & 1.70 & 3.60 & 0.40 & \underline{18.70} & \underline{8.70} & \textbf{9.30} \\
        \hline
        Ours & ICCV2025 & \textbf{46.50} & \textbf{17.46} & \textbf{64.60} & \underline{34.10} & \underline{32.00} & \textbf{12.50} & \textbf{26.90} & \textbf{26.70} & \textbf{7.50} & 2.90 & \textbf{3.30} & \underline{6.20} & \textbf{27.30} & \textbf{12.50} & \underline{28.00} & 2.30 & \textbf{5.10} & \underline{1.90} & \textbf{19.50} & \textbf{9.20} & \underline{9.20}\\
        \bottomrule
    \end{tabular}
    \caption{ Quantitative results on the hidden test set of SemanticKITTI~\cite{behley2019semantickitti}, where the highest and second-highest scores for each metric are highlighted in \textbf{bold} and \underline{underline}, respectively.} 
    \label{tab:smkitti}
\end{table*}

\textbf{Generating Occupancy Grid with 3D U-Net.} The resulting 3D feature is concatenated with refined context feature derived from VisFrontierNet ${F_{3D,2}}(v)$ and fed into a 3D U-Net to produce the final occupancy prediction. The 3D U-Net processes the modulated features and generates the occupancy grid for the entire set of voxels, effectively aligning the results with the image semantics while accounting for the noisy nature of the visible voxel predictions.

The denoising procedure can be summarized as follows.
\begin{equation}
O = \text{OcclusionMAE}(\overline{O}_V, t)
\end{equation}
By introducing a denoising procedure and injecting the noise level into the network, \emph{OcclusionMAE} learns to generate accurate occupancy predictions.

\subsection{Training Losses}
\label{loss}
To effectively train VisFrontierNet and OcclusionMAE, we employ a combination of several loss functions that capture different aspects of the 3D reconstruction problem. The primary loss is a category frequency-weighted cross-entropy loss $\mathcal{L}_{ce}$, which guides the learning of both networks. Additionally, we introduce the Scene-Class Affinity Loss, inspired by MonoScene~\cite{cao2022monoscene}, to separately constrain the recall, precision, and specificity of the scene's geometry and semantics, denoted as $\mathcal{L}_{geo}$ and $\mathcal{L}_{sem}$, respectively. Finally, a depth loss $\mathcal{L}_{d}$ is applied to ensure the coincidence of the probabilistic depth distribution $D_p$ and the depth map $D$.

\section{Experimental Evaluations}
In this section, we evaluate the performance of VisHall3D on two mainstream outdoor SSC (Semantic Scene Completion) datasets: SemanticKITTI~\cite{behley2019semantickitti} and SSCBench-KITTI-360~\cite{li2024sscbench,liao2022kitti}. We compare VisHall3D with state-of-the-art (SOTA) methods, including those utilizing stereo vision or temporal information. Furthermore, we conduct ablation studies to investigate the impact of each module on the model's performance.

\subsection{Dataset and Evaluation Metrics}
SemanticKITTI~\cite{behley2019semantickitti} and SSCBench-KITTI-360~\cite{li2024sscbench,liao2022kitti} are two widely-used outdoor SSC datasets. Both datasets adopt a scene range of $51.2\text{m}\times51.2\text{m}\times6.4\text{m}$, with each voxel having an edge length of $0.2\text{m}$, resulting in a scene resolution of $256\times256\times32$. SemanticKITTI consists of 10 training sequences (3,834 samples), 1 validation sequence (815 samples), and 11 test sequences (3,992 samples), with 20 valid classes and 1 invalid class. SSCBench-KITTI-360 provides 7 training sequences (8,487 samples), 1 validation sequence (1,812 samples), and 1 test sequence (2,566 samples), with 19 valid classes.

Following previous methods\cite{li2023voxformer,zheng2024monoocc,wang2024h2gformer,wang2024not,jiang2024symphonize,CGFormer}, we use MobileStereoNet\cite{shamsafar2022mobilestereonet} for depth prediction.

\definecolor{car}{RGB}{100, 150, 245}
\definecolor{bicycle}{RGB}{100, 230, 245}
\definecolor{motorcycle}{RGB}{30, 60, 150}
\definecolor{truck}{RGB}{80, 30, 180}
\definecolor{other-veh.}{RGB}{100, 80, 250}
\definecolor{person}{RGB}{255, 30, 30}
\definecolor{road}{RGB}{255, 0, 255}
\definecolor{parking}{RGB}{255, 150, 255}
\definecolor{sidewalk}{RGB}{75, 0, 75}
\definecolor{other-grnd.}{RGB}{175, 0, 75}
\definecolor{building}{RGB}{255, 200, 0}
\definecolor{fence}{RGB}{255, 120, 50}
\definecolor{vegetation}{RGB}{0, 175, 0}
\definecolor{terrain}{RGB}{150, 240, 80}
\definecolor{pole}{RGB}{255, 240, 150}
\definecolor{traf.-sign}{RGB}{255, 0, 0}
\definecolor{other-struct.}{RGB}{255, 120, 40}
\definecolor{other-obj.}{RGB}{100, 235, 255}

\begin{table*}[htbp]
\centering
\footnotesize
\setlength{\tabcolsep}{1pt}
\renewcommand{\arraystretch}{1.2}

\begin{tabular}{c|c|cc|cccccccccccccccccc}
 
\toprule
\textbf{Method} & \textbf{Date} & \textbf{IoU$\uparrow$} & \textbf{mIoU$\uparrow$} & \colorboxsquare{car}{car} & \colorboxsquare{bicycle}{bicycle} & \colorboxsquare{motorcycle}{motorcycle} & \colorboxsquare{truck}{truck} & \colorboxsquare{other-veh.}{other-veh.} & \colorboxsquare{person}{person} & \colorboxsquare{road}{road} & \colorboxsquare{parking}{parking} & \colorboxsquare{sidewalk}{sidewalk} & \colorboxsquare{other-grnd.}{other-grnd.} & \colorboxsquare{building}{building} & \colorboxsquare{fence}{fence} & \colorboxsquare{vegetation}{vegetation} & \colorboxsquare{terrain}{terrain} & \colorboxsquare{pole}{pole} & \colorboxsquare{traf.-sign}{traf.-sign} & \colorboxsquare{other-struct.}{other-struct.} & \colorboxsquare{other-obj.}{other-obj.} \\
\midrule
\multicolumn{21}{@{}l}{\quad\textbf{LiDAR-based methods}} \\ \hline
SSCNet~\cite{song2017semantic} & CVPR2017 & 53.58 & 16.95 & 31.95 & 0.00 & 0.17 & 10.29 & 0.00 & 0.07 & 65.70 & 17.33 & 41.24 & 3.22 & 44.41 & 6.77 & 43.72 & 28.87 & 0.78 & 0.75 & 8.69 & 0.67 \\
LMSCNet~\cite{roldao2020lmscnet}&3DV\phantom{R}2020& 47.35 & 13.65 & 20.91 & 0.00 & 0.00 & 0.26 & 0.58 & 0.00 & 62.95 & 13.51 & 33.51 & 0.20 & 43.67 & 0.33 & 40.01 & 26.80 & 0.00 & 0.00 & 3.63 & 0.00 \\
\hline

\multicolumn{21}{@{}l}{\quad\textbf{Monocular camera-based methods}} \\ \hline
\makebox[2.5cm][c]{MonoScene~\cite{cao2022monoscene}} &CVPR2023& 37.87  & 12.31 & 19.34 & 0.43 & 0.58 & 8.02 & 2.03 & 0.86 & 48.35 & 11.38 & 28.13 & 3.32 & 32.89 & 3.53 & 26.15 & 16.75 & 6.92 & 5.67 & 4.20 & 3.09\\
\makebox[2.5cm][c]{TPVFormer~\cite{huang2023tri}} &CVPR2023& 40.22 & 13.64 & 21.56 & 1.09 & 1.37 & 8.06 & 2.57 & 2.38 & 52.99 & 11.99 & 31.07 & 3.78 & 34.83 & 4.80 & 30.08 & 17.52 & 7.46 & 5.86 & 5.48 & 2.70 \\
\makebox[2.5cm][c]{OccFormer~\cite{zhang2023occformer}} &ICCV2023& 40.27 & 13.81 & 22.58 & 0.66 & 0.26 & 9.89 & 3.82 & 2.77 & 54.30 & 13.44 & 31.53 & 3.55 & 36.42 & 4.80 & 31.00 & 19.51 & 7.77 & 8.51 & 6.95 & 4.60 \\
\makebox[2.5cm][c]{VoxFormer~\cite{li2023voxformer}} &CVPR2023& 38.76 & 11.91 & 17.84 & 1.16 & 0.89 & 4.56 & 2.06 & 1.63 & 47.01 & 9.67 & 27.21 & 2.89 & 31.38 & 4.97 & 28.99 & 14.69 & 6.51 & 6.92 & 3.79 & 2.43 \\
\makebox[2.5cm][c]{IAMSSC~\cite{xiao2024instance}} &T-ITS2024& 41.80 & 12.97 & 18.53 & \underline{2.45} & 1.76 & 5.12 & 3.92 & 3.09 & 47.55 & 10.56 & 28.35 & 4.12 & 31.53 & 6.28 & 29.17 & 15.24 & 8.29 & 7.01 & 6.35 & 4.19 \\
\makebox[2.5cm][c]{DepthSSC~\cite{yao2023depthssc}} &arXiV2024& 40.85 & 14.28 & 21.90 & 2.36 & 4.30 & 11.51 & 4.56 & 2.92 & 50.88 & 12.89 & 30.27 & 2.49 & 37.33 & 5.22 & 29.61 & 21.59 & 5.97 & 7.71 & 5.24 & 3.51 \\
\makebox[2.5cm][c]{Symphonies~\cite{jiang2024symphonize}} &CVPR2024& 44.12 & 18.58 & \underline{30.02} & 1.85 & \underline{5.90} & \textbf{25.07} & \textbf{12.06} & \underline{8.20} & 54.94 & 13.83 & 32.76 & \textbf{6.93} & 35.11 & \underline{8.58} & 38.33 & 11.52 & 14.01 & 9.57 & \textbf{14.44} & \textbf{11.28}\\

CGFormer\cite{CGFormer}  &NIPS2024& \underline{48.07} & \underline{20.05} & 29.85 & \textbf{3.42} & 3.96 & 17.59 & 6.70 & 6.63 & \underline{63.85} & \underline{17.15} & \underline{40.72} & \underline{5.53} & \underline{42.73} & 8.22 & \underline{38.80} & \underline{24.04} & \underline{16.24} &  \underline{17.45} & 10.18 & 6.77\\
\hline
Ours&ICCV2025& \textbf{49.12} & \textbf{20.95} &\textbf{ 30.77} & 1.91 & \textbf{6.60} & \underline{17.99} & \underline{8.72} & \textbf{8.67} & \textbf{64.35} & \textbf{18.83} & \textbf{41.53} & 4.48 & \textbf{43.87} & \textbf{9.07} & \textbf{39.75} & \textbf{24.94} & \textbf{16.52} &  \textbf{20.66} &  \underline{10.30} & \underline{7.99}\\

\bottomrule
\end{tabular}
\caption{ Quantitative results on the test set of SSCBench-KITTI-360~\cite{li2024sscbench,liao2022kitti}, where the highest and second-highest scores for each metric are highlighted in \textbf{bold} and \underline{underline}, respectively.}
\label{tab:kitti360}
\end{table*}
\begin{table}
  \centering
  \begin{tabular}{@{}lccc@{}}
    \toprule
    \textbf{Method} & \textbf{Params(M)$\downarrow$} & \textbf{Memory(M)$\downarrow$} &  \textbf{Times(s)$\downarrow$}  \\
    \midrule
    Monoscene\cite{cao2022monoscene} & 149.5 & 19,041 & 0.49 \\
    CGFormer\cite{CGFormer} & 122.4 & 19,330 & 0.41 \\
    Ours & 127.8 & 22,597 & 0.34 \\
    \bottomrule
  \end{tabular}
  \vspace{-5pt}
  \caption{Computational Complexity and Memory Usage}
  \vspace{-15pt}
  \label{tab:memory}
\end{table}
\subsection{Comparisons}
We present the comparison of VisHall3D with SOTA methods on the SemanticKITTI~\cite{behley2019semantickitti} and SSCBench-KITTI-360~\cite{li2024sscbench,liao2022kitti} datasets in Tab.\ref{tab:smkitti} and Tab.\ref{tab:kitti360}, respectively. 

On the SemanticKITTI dataset, VisHall3D achieves an mIoU of 17.46\%, outperforming all existing monocular methods. Notably, VisHall3D surpasses the latest stereo method StereoScene~\cite{li2023bridging}, as well as the top-performing temporal method HTCL~\cite{li2024hierarchical} in terms of both IoU and mIoU. This demonstrates the effectiveness of our decoupled two-stage framework in capturing both the visible and invisible regions of the scene.

On the SSCBench-KITTI-360 dataset, VisHall3D sets a new state-of-the-art with an mIoU of 20.95\%, outperforming the previous best method CGFormer~\cite{CGFormer} by 0.90\%. Remarkably, VisHall3D even surpasses early LiDAR-based methods such as SCSCNet~\cite{song2017semantic} and LMSCNet~\cite{roldao2020lmscnet}, showcasing the potential of monocular methods in capturing complex 3D scenes.

Fig.~\ref{fig:visualization} presents a qualitative comparison of VisHall3D with MonoScene~\cite{cao2022monoscene} and the ground truth on the SemanticKITTI dataset. VisHall3D generates more accurate and complete scene reconstructions, especially in the invisible regions occluded by foreground objects. This visual comparison further validates the superiority of our method in capturing the global scene context and hallucinating plausible geometries.

We also compare VisHall3D against SOTA methods in terms of computational complexity in Tab.~\ref{tab:memory}. Despite our two-stage framework, VisHall3D maintains parameters comparable to CGFormer, while outperforming it in mIoU and IoU. The slight memory increase is justified by performance gains. Under identical conditions, VisHall3D runs faster than CGFormer and Monoscene while delivering superior reconstruction quality.

\subsection{Ablation Studies}
\label{abl}
In line with other works~\cite{cao2022monoscene,li2023voxformer,jiang2024symphonize,CGFormer}, our ablation studies are primarily conducted on the SemanticKITTI~\cite{behley2019semantickitti} validation set. These studies encompass three key aspects: (1) the overall architectural components, (2) the division of visible and invisible regions, and (3) the noise incorporated in the OcclusionMAE.

\begin{table*}[htbp] 
\centering
\begin{tabular}{lcccc}
\toprule
\textbf{Method} & \textbf{IoU$\uparrow$} & \textbf{mIoU$\uparrow$} & \textbf{Params (M)} & \textbf{Memory (M)} \\
\midrule
Baseline & 42.11 & 14.56 & 57.2 & 15260 \\
VisFrontierNet w/o Feature Refinement & \textbf{46.44} \textcolor{red}{(+4.33)} & 15.11 \textcolor{red}{(+0.55)} & 74.3 & 17349 \\
\midrule
+ OcclusionMAE w/o Denoising  & 46.38 \textcolor{red}{(-0.06)} & 15.99 \textcolor{red}{(+0.88)} & 86.7 & 18746 \\
+ Feature Refinement & 45.88 \textcolor{red}{(-0.50)} & 16.59 \textcolor{red}{(+0.60)} & 125.5 & 21246 \\
+ Denoising & 46.14 \textcolor{red}{(+0.26)} & \textbf{17.06} \textcolor{red}{(+0.47)} & 127.8 & 22597 \\
\bottomrule
\end{tabular}
\caption{Ablation study of the architectural components on the validation set of SemanticKITTI~\cite{behley2019semantickitti}, where the model initially used the hard assignment method from \cite{li2023voxformer,jiang2024symphonize} for lifting 2D features to 3D.}
\label{tab:archi}
\end{table*}

\textbf{Overall Architectural Components.} Tab.~\ref{tab:archi} presents our component-wise analysis of the network architecture. The baseline model retains the backbone and neck but adopts a direct hard-assign lifting approach from~\cite{li2023voxformer,jiang2024symphonize}, where each pixel corresponds to a single voxel. Its architecture can be considered as a variation of Symphonies~\cite{jiang2024symphonize} where the Symphonies Decoder is replaced with a 3D UNet.

The introduction of unsigned distance function for the visual frontier not only enhances the mIoU but also significantly improves the IoU by 4.33\%, likely due to its superior preservation of geometric information for the visual frontier in the predicted depth map. The unsigned distance function provides a more informative representation of the visual frontier compared to binary masks, enabling the network to better capture the visibility relationships between voxels. Subsequently, by incorporating the OcclusionMAE, we divided the network into two parts: the first part maintains the original structure but predicts voxels for visible regions only, while the second part, the OcclusionMAE, hallucinates the entire scene, including invisible regions. Although this does not improve the IoU, it substantially boosts the mIoU by 0.88\%.

Further, we integrated the feature refinement based on 3D deformable attention to enhance the model's perception of visible regions. While this results in a slight decrease in IoU by 0.50\%, the mIoU continues to improve by 0.60\%, reaching 16.59. Finally, to enhance the capability of the OcclusionMAE, we introduced a denoising strategy, which further improves the mIoU by 0.47\%.


Interestingly, VisFrontierNet without feature refinement achieves the highest IoU, showing that its unsigned distance function alone offers a strong geometric prior for accurate visible region prediction.

\begin{table}[htbp] 
\centering
\begin{tabular}{cccc}
\toprule
\textbf{Invisibility} & \textbf{Threshold $\theta_d$ (m)} & \textbf{IoU$\uparrow$} & \textbf{mIoU$\uparrow$} \\
\midrule
OOV & - & 43.98 & 15.89 \\
\midrule
OOV + Occ.  & 1.5 & 45.87 & 16.65 \\
OOV + Occ.  & 2.5  & 45.83 &16.92  \\
OOV + Occ. & 3.5  & \textbf{46.14}   &\textbf{17.06} \\
OOV + Occ. & 4.5  & 45.94   & 16.95 \\
\bottomrule
\end{tabular}
\caption{Ablation study on visible and invisible region division on SemanticKITTI~\cite{behley2019semantickitti}'s validation set, where OOV and Occ. refer to using the out of view  and occlusion respectively.}
\label{tab:visibly}
\end{table} 

\textbf{Visibility Identification.} Decoupling visible and invisible scenes is one of our most critical components, and the key lies in how to delineate what constitutes the visible and invisible regions. Tab.~\ref{tab:visibly} presents our ablation study on the division of visible and invisible regions.

The baseline approach solely relies on the out of view checking to distinguish visible from invisible areas, considering only whether voxels project onto the image plane while entirely ignoring occlusion issues, As shown in Tab.~\ref{tab:visibly}, this naive implementation performs significantly worse than methods that account for occlusion. In occlusion-aware methods, we fine-tuned the threshold and found that 3.5m gives the best results, suggesting moderate visibility relaxation helps handle occlusion and depth uncertainty. However, larger thresholds (e.g., 4.5m) introduce noise and harm performance.


\begin{table}[htbp] 
\centering
\begin{tabular}{cccc}
\toprule
\textbf{$R_h$ (Voxel)} & \textbf{$R_d$ (Voxel)} & \textbf{IoU$\uparrow$} & \textbf{mIoU$\uparrow$}\\
\midrule
0 & 0 & 46.03 & 16.32 \\
\hline
0 & 2 & 45.92&  16.79 \\
0 & 3 & 45.83 & \textbf{16.92} \\
0 & 4 & 45.73 & 16.66  \\
\hline
1  & 3 & \textbf{46.19} &16.52  \\
1 &  4 & 46.12 & 16.69\\
\bottomrule
\end{tabular}
\caption{Ablation study on the noise level in OcclusionMAE on SemanticKITTI~\cite{behley2019semantickitti}, where $R_h$ and $R_d$ refer to the noise range in the horizontal and depth directions, respectively (conducted with the threshold $\theta_d$ of 2.5m).}
\label{tab:noise}
\end{table}
\textbf{Noise in the OcclusionMAE.} To achieve effective hallucination, we introduce a certain level of noise into the OcclusionMAE, which also enhances the network's robustness. Tab.~\ref{tab:noise} presents our study on the noise parameters in the Hallucinating MAE. These parameters consist of two components: the maximum sampling distance in the horizontal direction and the depth direction. When no noise is applied (both parameters are set to 0), the model performs worst on mIoU. In contrast, the best mIoU is achieved when the horizontal noise is set to 0 and the depth noise to 3.

Interestingly, the impact of increasing horizontal noise on the network is significantly greater than that of depth noise, likely due to the inherent inaccuracies in depth estimation. This finding suggests that the model is more sensitive to the perturbation in the horizontal direction, and a careful balance between the two noise components is crucial for optimal performance.

\section{Conclusion}
In this paper, we propose VisHall3D, a two-stage framework for monocular semantic scene completion that tackles feature entanglement and geometric inconsistency by explicitly separating visible-region reconstruction and invisible-region hallucination, and demonstrate its effectiveness with extensive experiments.

\section*{Acknowledgements}
This paper is supported by the National Natural Science Foundation of China (No.62473306, N0.U24B20181, No.62441616) and Science and Technology Research and Development Plan of China State Railway Group Co., Ltd. (No. RITS2023KF03).

{
    \small
    \bibliographystyle{ieeenat_fullname}
    \bibliography{main}

\begin{thebibliography}{49}
\providecommand{\natexlab}[1]{#1}
\providecommand{\url}[1]{\texttt{#1}}
\expandafter\ifx\csname urlstyle\endcsname\relax
  \providecommand{\doi}[1]{doi: #1}\else
  \providecommand{\doi}{doi: \begingroup \urlstyle{rm}\Url}\fi

\bibitem[Behley et~al.(2019)Behley, Garbade, Milioto, Quenzel, Behnke, Stachniss, and Gall]{behley2019semantickitti}
Jens Behley, Martin Garbade, Andres Milioto, Jan Quenzel, Sven Behnke, Cyrill Stachniss, and Jurgen Gall.
\newblock Semantickitti: A dataset for semantic scene understanding of lidar sequences.
\newblock In \emph{Proceedings of the IEEE/CVF international conference on computer vision}, pages 9297--9307, 2019.

\bibitem[Cai and Vasconcelos(2019)]{cai2019cascade}
Zhaowei Cai and Nuno Vasconcelos.
\newblock Cascade r-cnn: High quality object detection and instance segmentation.
\newblock \emph{IEEE transactions on pattern analysis and machine intelligence}, 43\penalty0 (5):\penalty0 1483--1498, 2019.

\bibitem[Cao and De~Charette(2022)]{cao2022monoscene}
Anh-Quan Cao and Raoul De~Charette.
\newblock Monoscene: Monocular 3d semantic scene completion.
\newblock In \emph{Proceedings of the IEEE/CVF Conference on Computer Vision and Pattern Recognition}, pages 3991--4001, 2022.

\bibitem[Carion et~al.(2020)Carion, Massa, Synnaeve, Usunier, Kirillov, and Zagoruyko]{carion2020end}
Nicolas Carion, Francisco Massa, Gabriel Synnaeve, Nicolas Usunier, Alexander Kirillov, and Sergey Zagoruyko.
\newblock End-to-end object detection with transformers.
\newblock In \emph{European conference on computer vision}, pages 213--229. Springer, 2020.

\bibitem[Chen et~al.(2022)Chen, Wang, Wang, Tian, Xiong, and Li]{chen2022epro}
Hansheng Chen, Pichao Wang, Fan Wang, Wei Tian, Lu Xiong, and Hao Li.
\newblock Epro-pnp: Generalized end-to-end probabilistic perspective-n-points for monocular object pose estimation.
\newblock In \emph{Proceedings of the IEEE/CVF conference on computer vision and pattern recognition}, pages 2781--2790, 2022.

\bibitem[Cheng et~al.(2021)Cheng, Agia, Ren, Li, and Bingbing]{cheng2021s3cnet}
Ran Cheng, Christopher Agia, Yuan Ren, Xinhai Li, and Liu Bingbing.
\newblock S3cnet: A sparse semantic scene completion network for lidar point clouds.
\newblock In \emph{Conference on Robot Learning}, pages 2148--2161. PMLR, 2021.

\bibitem[Girshick(2015)]{girshick2015fast}
Ross Girshick.
\newblock Fast r-cnn.
\newblock In \emph{Proceedings of the IEEE international conference on computer vision}, pages 1440--1448, 2015.

\bibitem[He et~al.(2016)He, Zhang, Ren, and Sun]{he2016deep}
Kaiming He, Xiangyu Zhang, Shaoqing Ren, and Jian Sun.
\newblock Deep residual learning for image recognition.
\newblock In \emph{Proceedings of the IEEE conference on computer vision and pattern recognition}, pages 770--778, 2016.

\bibitem[He et~al.(2022)He, Chen, Xie, Li, Doll{\'a}r, and Girshick]{he2022masked}
Kaiming He, Xinlei Chen, Saining Xie, Yanghao Li, Piotr Doll{\'a}r, and Ross Girshick.
\newblock Masked autoencoders are scalable vision learners.
\newblock In \emph{Proceedings of the IEEE/CVF conference on computer vision and pattern recognition}, pages 16000--16009, 2022.

\bibitem[Huang and Belongie(2017)]{huang2017arbitrary}
Xun Huang and Serge Belongie.
\newblock Arbitrary style transfer in real-time with adaptive instance normalization.
\newblock In \emph{Proceedings of the IEEE international conference on computer vision}, pages 1501--1510, 2017.

\bibitem[Huang et~al.(2023)Huang, Zheng, Zhang, Zhou, and Lu]{huang2023tri}
Yuanhui Huang, Wenzhao Zheng, Yunpeng Zhang, Jie Zhou, and Jiwen Lu.
\newblock Tri-perspective view for vision-based 3d semantic occupancy prediction.
\newblock In \emph{Proceedings of the IEEE/CVF conference on computer vision and pattern recognition}, pages 9223--9232, 2023.

\bibitem[Jiang et~al.(2024)Jiang, Cheng, Gao, Zhang, Lin, Liu, and Wang]{jiang2024symphonize}
Haoyi Jiang, Tianheng Cheng, Naiyu Gao, Haoyang Zhang, Tianwei Lin, Wenyu Liu, and Xinggang Wang.
\newblock Symphonize 3d semantic scene completion with contextual instance queries.
\newblock In \emph{Proceedings of the IEEE/CVF Conference on Computer Vision and Pattern Recognition}, pages 20258--20267, 2024.

\bibitem[Lavreniuk(2024)]{lavreniuk2024spidepth}
Mykola Lavreniuk.
\newblock Spidepth: Strengthened pose information for self-supervised monocular depth estimation.
\newblock \emph{arXiv preprint arXiv:2404.12501}, 2024.

\bibitem[Li et~al.(2023{\natexlab{a}})Li, Sun, Liang, Du, Zhang, Wang, Wang, Jin, and Zeng]{li2023bridging}
Bohan Li, Yasheng Sun, Zhujin Liang, Dalong Du, Zhuanghui Zhang, Xiaofeng Wang, Yunnan Wang, Xin Jin, and Wenjun Zeng.
\newblock Bridging stereo geometry and bev representation with reliable mutual interaction for semantic scene completion.
\newblock \emph{arXiv preprint arXiv:2303.13959}, 2023{\natexlab{a}}.

\bibitem[Li et~al.(2024{\natexlab{a}})Li, Deng, Zhang, Liang, Du, Jin, and Zeng]{li2024hierarchical}
Bohan Li, Jiajun Deng, Wenyao Zhang, Zhujin Liang, Dalong Du, Xin Jin, and Wenjun Zeng.
\newblock Hierarchical temporal context learning for camera-based semantic scene completion.
\newblock \emph{arXiv preprint arXiv:2407.02077}, 2024{\natexlab{a}}.

\bibitem[Li et~al.(2023{\natexlab{b}})Li, Zhang, Xu, Liu, Zhang, Ni, and Shum]{li2023mask}
Feng Li, Hao Zhang, Huaizhe Xu, Shilong Liu, Lei Zhang, Lionel~M Ni, and Heung-Yeung Shum.
\newblock Mask dino: Towards a unified transformer-based framework for object detection and segmentation.
\newblock In \emph{Proceedings of the IEEE/CVF conference on computer vision and pattern recognition}, pages 3041--3050, 2023{\natexlab{b}}.

\bibitem[Li et~al.(2023{\natexlab{c}})Li, Zhang, Zeng, Liu, Li, Ren, and Zhang]{li2023dfa3d}
Hongyang Li, Hao Zhang, Zhaoyang Zeng, Shilong Liu, Feng Li, Tianhe Ren, and Lei Zhang.
\newblock Dfa3d: 3d deformable attention for 2d-to-3d feature lifting.
\newblock In \emph{Proceedings of the IEEE/CVF International Conference on Computer Vision}, pages 6684--6693, 2023{\natexlab{c}}.

\bibitem[Li et~al.(2019)Li, Liu, Yuan, Zhao, Siegwart, Reid, and Cadena]{li2019depth}
Jie Li, Yu Liu, Xia Yuan, Chunxia Zhao, Roland Siegwart, Ian Reid, and Cesar Cadena.
\newblock Depth based semantic scene completion with position importance aware loss.
\newblock \emph{IEEE Robotics and Automation Letters}, 5\penalty0 (1):\penalty0 219--226, 2019.

\bibitem[Li et~al.(2020)Li, Han, Wang, Liu, and Yuan]{li2020anisotropic}
Jie Li, Kai Han, Peng Wang, Yu Liu, and Xia Yuan.
\newblock Anisotropic convolutional networks for 3d semantic scene completion.
\newblock In \emph{Proceedings of the IEEE/CVF Conference on Computer Vision and Pattern Recognition}, pages 3351--3359, 2020.

\bibitem[Li et~al.(2023{\natexlab{d}})Li, Yu, Choy, Xiao, Alvarez, Fidler, Feng, and Anandkumar]{li2023voxformer}
Yiming Li, Zhiding Yu, Christopher Choy, Chaowei Xiao, Jose~M Alvarez, Sanja Fidler, Chen Feng, and Anima Anandkumar.
\newblock Voxformer: Sparse voxel transformer for camera-based 3d semantic scene completion.
\newblock In \emph{Proceedings of the IEEE/CVF conference on computer vision and pattern recognition}, pages 9087--9098, 2023{\natexlab{d}}.

\bibitem[Li et~al.(2024{\natexlab{b}})Li, Li, Liu, Gong, Li, Chen, Wang, Li, Jiang, Yu, Wang, Zhao, Yu, and Feng]{li2024sscbench}
Yiming Li, Sihang Li, Xinhao Liu, Moonjun Gong, Kenan Li, Nuo Chen, Zijun Wang, Zhiheng Li, Tao Jiang, Fisher Yu, Yue Wang, Hang Zhao, Zhiding Yu, and Chen Feng.
\newblock Sscbench: A large-scale 3d semantic scene completion benchmark for autonomous driving.
\newblock In \emph{2024 IEEE/RSJ International Conference on Intelligent Robots and Systems (IROS)}, 2024{\natexlab{b}}.

\bibitem[Li et~al.()Li, Wang, Li, Xie, Sima, Lu, Qiao, and Dai]{li2022bevformer}
Zhiqi Li, Wenhai Wang, Hongyang Li, Enze Xie, Chonghao Sima, Tong Lu, Yu Qiao, and Jifeng Dai.
\newblock Bevformer: Learning bird’s-eye-view representation from multi-camera images via spatiotemporal transformers.

\bibitem[Li et~al.(2022)Li, Qu, Zhou, Liu, Wang, and Jiang]{li2022diversity}
Zhuoling Li, Zhan Qu, Yang Zhou, Jianzhuang Liu, Haoqian Wang, and Lihui Jiang.
\newblock Diversity matters: Fully exploiting depth clues for reliable monocular 3d object detection.
\newblock In \emph{Proceedings of the IEEE/CVF Conference on Computer Vision and Pattern Recognition}, pages 2791--2800, 2022.

\bibitem[Li et~al.(2023{\natexlab{e}})Li, Yu, Austin, Fang, Lan, Kautz, and Alvarez]{li2023fb}
Zhiqi Li, Zhiding Yu, David Austin, Mingsheng Fang, Shiyi Lan, Jan Kautz, and Jose~M Alvarez.
\newblock Fb-occ: 3d occupancy prediction based on forward-backward view transformation.
\newblock \emph{arXiv preprint arXiv:2307.01492}, 2023{\natexlab{e}}.

\bibitem[Liao et~al.(2022)Liao, Xie, and Geiger]{liao2022kitti}
Yiyi Liao, Jun Xie, and Andreas Geiger.
\newblock Kitti-360: A novel dataset and benchmarks for urban scene understanding in 2d and 3d.
\newblock \emph{IEEE Transactions on Pattern Analysis and Machine Intelligence}, 45\penalty0 (3):\penalty0 3292--3310, 2022.

\bibitem[Liu et~al.(2024)Liu, Zhang, Kong, Zhang, Wu, Ding, Xu, Ming, Wei, and Liu]{liu2024occtransformer}
Jian Liu, Sipeng Zhang, Chuixin Kong, Wenyuan Zhang, Yuhang Wu, Yikang Ding, Borun Xu, Ruibo Ming, Donglai Wei, and Xianming Liu.
\newblock Occtransformer: Improving bevformer for 3d camera-only occupancy prediction.
\newblock \emph{arXiv preprint arXiv:2402.18140}, 2024.

\bibitem[Liu et~al.(2021)Liu, Zhou, Lu, Fang, and Zhang]{liu2021autoshape}
Zongdai Liu, Dingfu Zhou, Feixiang Lu, Jin Fang, and Liangjun Zhang.
\newblock Autoshape: Real-time shape-aware monocular 3d object detection.
\newblock In \emph{Proceedings of the IEEE/CVF International Conference on Computer Vision}, pages 15641--15650, 2021.

\bibitem[Luo et~al.(2021)Luo, Dai, Shao, and Ding]{luo2021m3dssd}
Shujie Luo, Hang Dai, Ling Shao, and Yong Ding.
\newblock M3dssd: Monocular 3d single stage object detector.
\newblock In \emph{Proceedings of the IEEE/CVF conference on computer vision and pattern recognition}, pages 6145--6154, 2021.

\bibitem[Miao et~al.(2023)Miao, Liu, Chen, Gong, Xu, Hu, and Zhou]{miao2023occdepth}
Ruihang Miao, Weizhou Liu, Mingrui Chen, Zheng Gong, Weixin Xu, Chen Hu, and Shuchang Zhou.
\newblock Occdepth: A depth-aware method for 3d semantic scene completion.
\newblock \emph{arXiv preprint arXiv:2302.13540}, 2023.

\bibitem[Philion and Fidler(2020)]{philion2020lift}
Jonah Philion and Sanja Fidler.
\newblock Lift, splat, shoot: Encoding images from arbitrary camera rigs by implicitly unprojecting to 3d.
\newblock In \emph{Proceedings of the European Conference on Computer Vision}, 2020.

\bibitem[Rist et~al.(2021)Rist, Emmerichs, Enzweiler, and Gavrila]{rist2021semantic}
Christoph~B Rist, David Emmerichs, Markus Enzweiler, and Dariu~M Gavrila.
\newblock Semantic scene completion using local deep implicit functions on lidar data.
\newblock \emph{IEEE transactions on pattern analysis and machine intelligence}, 44\penalty0 (10):\penalty0 7205--7218, 2021.

\bibitem[Roldao et~al.(2020)Roldao, De~Charette, and Verroust-Blondet]{roldao2020lmscnet}
Luis Roldao, Raoul De~Charette, and Anne Verroust-Blondet.
\newblock Lmscnet: Lightweight multiscale 3d semantic completion.
\newblock In \emph{2020 International Conference on 3D Vision (3DV)}, pages 111--119. IEEE, 2020.

\bibitem[Shamsafar et~al.(2022)Shamsafar, Woerz, Rahim, and Zell]{shamsafar2022mobilestereonet}
Faranak Shamsafar, Samuel Woerz, Rafia Rahim, and Andreas Zell.
\newblock Mobilestereonet: Towards lightweight deep networks for stereo matching.
\newblock In \emph{Proceedings of the ieee/cvf winter conference on applications of computer vision}, pages 2417--2426, 2022.

\bibitem[Song et~al.(2017)Song, Yu, Zeng, Chang, Savva, and Funkhouser]{song2017semantic}
Shuran Song, Fisher Yu, Andy Zeng, Angel~X Chang, Manolis Savva, and Thomas Funkhouser.
\newblock Semantic scene completion from a single depth image.
\newblock In \emph{Proceedings of the IEEE conference on computer vision and pattern recognition}, pages 1746--1754, 2017.

\bibitem[Sun et~al.(2021)Sun, Zhang, Jiang, Kong, Xu, Zhan, Tomizuka, Li, Yuan, Wang, et~al.]{sun2021sparse}
Peize Sun, Rufeng Zhang, Yi Jiang, Tao Kong, Chenfeng Xu, Wei Zhan, Masayoshi Tomizuka, Lei Li, Zehuan Yuan, Changhu Wang, et~al.
\newblock Sparse r-cnn: End-to-end object detection with learnable proposals.
\newblock In \emph{Proceedings of the IEEE/CVF conference on computer vision and pattern recognition}, pages 14454--14463, 2021.

\bibitem[Wang et~al.(2024)Wang, Yu, Li, Liu, Liu, Chen, and Zhu]{wang2024not}
Song Wang, Jiawei Yu, Wentong Li, Wenyu Liu, Xiaolu Liu, Junbo Chen, and Jianke Zhu.
\newblock Not all voxels are equal: Hardness-aware semantic scene completion with self-distillation.
\newblock In \emph{Proceedings of the IEEE/CVF Conference on Computer Vision and Pattern Recognition}, pages 14792--14801, 2024.

\bibitem[Wang and Tong(2024)]{wang2024h2gformer}
Yu Wang and Chao Tong.
\newblock H2gformer: Horizontal-to-global voxel transformer for 3d semantic scene completion.
\newblock In \emph{Proceedings of the AAAI Conference on Artificial Intelligence}, pages 5722--5730, 2024.

\bibitem[Wei et~al.(2023)Wei, Zhao, Zheng, Zhu, Zhou, and Lu]{wei2023surroundocc}
Yi Wei, Linqing Zhao, Wenzhao Zheng, Zheng Zhu, Jie Zhou, and Jiwen Lu.
\newblock Surroundocc: Multi-camera 3d occupancy prediction for autonomous driving.
\newblock In \emph{Proceedings of the IEEE/CVF International Conference on Computer Vision}, pages 21729--21740, 2023.

\bibitem[Xiao et~al.(2024)Xiao, Xu, Kang, and Li]{xiao2024instance}
Haihong Xiao, Hongbin Xu, Wenxiong Kang, and Yuqiong Li.
\newblock Instance-aware monocular 3d semantic scene completion.
\newblock \emph{IEEE Transactions on Intelligent Transportation Systems}, 25\penalty0 (7):\penalty0 6543--6554, 2024.

\bibitem[Yan et~al.(2021)Yan, Gao, Li, Zhang, Li, Huang, and Cui]{yan2021sparse}
Xu Yan, Jiantao Gao, Jie Li, Ruimao Zhang, Zhen Li, Rui Huang, and Shuguang Cui.
\newblock Sparse single sweep lidar point cloud segmentation via learning contextual shape priors from scene completion.
\newblock In \emph{Proceedings of the AAAI conference on artificial intelligence}, pages 3101--3109, 2021.

\bibitem[Yao and Zhang(2023)]{yao2023depthssc}
Jiawei Yao and Jusheng Zhang.
\newblock Depthssc: Depth-spatial alignment and dynamic voxel resolution for monocular 3d semantic scene completion.
\newblock \emph{arXiv preprint arXiv:2311.17084}, 2023.

\bibitem[Yao et~al.(2023)Yao, Li, Sun, Cai, Li, Ouyang, and Li]{yao2023ndc}
Jiawei Yao, Chuming Li, Keqiang Sun, Yingjie Cai, Hao Li, Wanli Ouyang, and Hongsheng Li.
\newblock Ndc-scene: Boost monocular 3d semantic scene completion in normalized device coordinates space.
\newblock In \emph{2023 IEEE/CVF International Conference on Computer Vision (ICCV)}, pages 9421--9431. IEEE Computer Society, 2023.

\bibitem[Yu et~al.(2024)Yu, Zhang, Ying, Yu, Hu, Luo, Cao, and Shen]{CGFormer}
Zhu Yu, Runmin Zhang, Jiacheng Ying, Junchen Yu, Xiaohai Hu, Lun Luo, Si-Yuan Cao, and Hui-liang Shen.
\newblock Context and geometry aware voxel transformer for semantic scene completion.
\newblock In \emph{Advances in Neural Information Processing Systems}, pages 1531--1555, 2024.

\bibitem[Zhang et~al.(2023{\natexlab{a}})Zhang, Qiu, Wang, Guo, Cui, Qiao, Li, and Gao]{zhang2023monodetr}
Renrui Zhang, Han Qiu, Tai Wang, Ziyu Guo, Ziteng Cui, Yu Qiao, Hongsheng Li, and Peng Gao.
\newblock Monodetr: Depth-guided transformer for monocular 3d object detection.
\newblock In \emph{Proceedings of the IEEE/CVF International Conference on Computer Vision}, pages 9155--9166, 2023{\natexlab{a}}.

\bibitem[Zhang et~al.(2021)Zhang, Lu, and Zhou]{zhang2021objects}
Yunpeng Zhang, Jiwen Lu, and Jie Zhou.
\newblock Objects are different: Flexible monocular 3d object detection.
\newblock In \emph{Proceedings of the IEEE/CVF Conference on Computer Vision and Pattern Recognition}, pages 3289--3298, 2021.

\bibitem[Zhang et~al.(2023{\natexlab{b}})Zhang, Zhu, and Du]{zhang2023occformer}
Yunpeng Zhang, Zheng Zhu, and Dalong Du.
\newblock Occformer: Dual-path transformer for vision-based 3d semantic occupancy prediction.
\newblock In \emph{Proceedings of the IEEE/CVF International Conference on Computer Vision}, pages 9433--9443, 2023{\natexlab{b}}.

\bibitem[Zheng et~al.(2024)Zheng, Li, Li, Zheng, Jin, Zhong, Long, Zhao, and Zhang]{zheng2024monoocc}
Yupeng Zheng, Xiang Li, Pengfei Li, Yuhang Zheng, Bu Jin, Chengliang Zhong, Xiaoxiao Long, Hao Zhao, and Qichao Zhang.
\newblock Monoocc: Digging into monocular semantic occupancy prediction.
\newblock In \emph{2024 IEEE International Conference on Robotics and Automation (ICRA)}, pages 18398--18405. IEEE, 2024.

\bibitem[Zhou et~al.(2019)Zhou, Wang, and Kr{\"a}henb{\"u}hl]{zhou2019objects}
Xingyi Zhou, Dequan Wang, and Philipp Kr{\"a}henb{\"u}hl.
\newblock Objects as points.
\newblock \emph{arXiv preprint arXiv:1904.07850}, 2019.

\bibitem[Zhu et~al.(2020)Zhu, Su, Lu, Li, Wang, and Dai]{zhu2020deformable}
Xizhou Zhu, Weijie Su, Lewei Lu, Bin Li, Xiaogang Wang, and Jifeng Dai.
\newblock Deformable detr: Deformable transformers for end-to-end object detection.
\newblock \emph{arXiv preprint arXiv:2010.04159}, 2020.

\end{thebibliography}
}

\end{document}